\documentclass[journal,onecolumn]{IEEEtran}
\usepackage[utf8]{inputenc}
\IEEEoverridecommandlockouts
\usepackage{cite}
\usepackage[justification=centering]{caption}
\usepackage{amsmath,amssymb,amsfonts}
\usepackage{tikz}
\usepackage{pgfplots}
\usepgfplotslibrary{groupplots}
\pgfplotsset{compat=1.18}
\usepackage{algorithm}
\usepackage{algorithmic}
\usepackage{graphicx}
\usepackage{textcomp}
\usepackage{xcolor}
\usepackage{booktabs}
\usepackage{multirow}
\usepackage{enumitem}
\usepackage{url}
\def\BibTeX{{\rm B\kern-.05em{\sc i\kern-.025em b}\kern-.08em
    T\kern-.1667em\lower.7ex\hbox{E}\kern-.125emX}}

\begin{document}

\title{A Novel Graph Fraud Detector via Grouped Attribute Completion and Confidence-Aware Contrastive Learning}

\author{\IEEEauthorblockN{1\textsuperscript{st} Junpeng Wu}

\IEEEauthorblockA{\textit{College of Computer and Information Science} \\
\textit{Southwest University}\\
Chongqing, China \\
junpengw67@gmail.com}
\and

\IEEEauthorblockN{Ye Yuan*}

\IEEEauthorblockA{\textit{College of Computer and Information Science} \\
\textit{Southwest University}\\
Chongqing, China \\
*yuanyekl@swu.edu.cn}
}

\maketitle

\begin{abstract}
Graph fraud detection plays a pivotal role in safeguarding the security and integrity of modern digital ecosystems. Graph Neural Networks (GNNs) are commonly adopted for graph fraud detection. However, the practical performance of existing GNN-based detectors is severely hindered by incomplete node attributes and extreme class imbalance within graphs. To mitigate these limitations, this paper proposes a novel framework for \underline{G}raph \underline{F}raud \underline{D}etection with \underline{G}rouped attribute completion and \underline{C}onfidence-aware \underline{C}ontrastive learning, named GFD-GC. Specifically, it first imitates heterogeneous neighborhood structures to implement group-wise aggregation, which obtains informative complete node features by capturing fine-grained graph contextual patterns. Further, it introduces a confidence-aware supervised contrastive learning strategy to augment scarce labeled fraud nodes with high-confidence pseudo-fraud nodes, which enhances the compactness of fraud representations and their separability from non-fraud nodes. Extensive experiments demonstrate the superiority of the proposed GFD-GC over state-of-the-art baselines on the graph fraud detection task, thereby providing an effective solution for real-world fraud scenarios.
\end{abstract}

\section{INTRODUCTION}
The rapid growth of digital ecosystems, such as online payments, e-commerce platforms, and social networks, has been accompanied by increasing fraudulent activities, causing substantial financial losses and undermining system reliability \cite{qinetheft2026,crypto2025}. In practice, fraud rarely occurs in isolation. Instead, it usually involves complex interactions among users, accounts, devices, merchants, and transactions. Such relations can be naturally modeled as graphs, where fraud detection is formulated as a node classification task on attributed graphs.

Motivated by this observation, graph neural networks (GNNs) have been widely used for graph fraud detection, due to their ability to capture structural dependency and neighborhood context \cite{gcn,graphsage,gat,gin,gtcn2026,ncigcn2025,mgcn2025,gta2t2025,highordergcn2026,sgddyg2025,deepwalk2014,node2vec2016,line2015,chebnet2016,sgc2019,appnp2019,jknet2018,vgae2016,dgmixer2025,stgnntensor2025,tccn2025,cbnlf2025,labc2025,mirna2025,mcdgnn2026,ncsac2026}. Existing graph-based methods have achieved promising results in applications such as electricity theft detection, review spam detection, financial risk control, transaction-network analysis, and social fraud analysis \cite{qinetheft2026,crypto2025,reviewnet,yelpfraud,caregnn,pcgnn,sgddyg2025,kdmil2025,fmvpci2025,sentiment2025,emotion2025,spsa2025,batterylife2025,kdaerial2024,robotarmsystem2024,anytime2024,fmoe2026}. However, their practical performance is still limited in real-world scenarios.

The first challenge is incomplete node attributes. Most GNN-based fraud detectors rely heavily on node features to characterize suspicious behaviors. When features are missing, the learned representations can become unreliable. Although prior studies handle missing attributes through graph-aware incomplete-data modeling or robust representation learning \cite{glcp2025,orae2025,mmlf2024,ekflfa2026,apidnlf2026,lerpid2025,pinlf2025,fuzzypid2024,adnlfa2024,kflfa2023,adnlfm2023,mrlfm2022,gfnlf2020,kfwebqos2020,nghfl2026,modetucker2025,dsrpso2026,tlroc2026,antf2025,gatwostep2025,mmae2025,hifreview2025,snrtlft2026,amnlft2025,biasnlt2025,masntf2025,nnlft2026,batterylife2025,rllfa2025,ntf2025,trafficimputation2025,fgreg2024,admmnlf2024,apsgd2024,gcnmf2021,sat2022,featureprop2022}, they are mainly designed for general representation learning and do not explicitly model the heterogeneous neighborhood patterns in fraud graphs.

The second challenge is extreme class imbalance. In fraud graphs, fraudulent nodes are much fewer than benign ones, making positive supervision highly insufficient. This often leads to weak fraud representations and unclear decision boundaries. Recent studies show that contrastive learning can improve representation robustness and separability \cite{supcon,qinetheft2026,dgi2019,graphcl2020,mvgrl2020,grace2020,gca2021,simclr2020,moco2020,graphsmote2021,dominant2019,anomalydae2020,cola2022}. However, directly applying supervised contrastive learning remains inadequate when labeled fraud nodes are extremely scarce.

To address the above challenges in a unified manner, we propose a novel framework for \underline{G}raph \underline{F}raud \underline{D}etection with \underline{G}rouped attribute completion and \underline{C}onfidence-aware \underline{C}ontrastive learning, termed \textbf{GFD-GC}. The key idea is to jointly improve node representation quality from both the feature-completion perspective and the representation-discrimination perspective. Specifically, GFD-GC first performs group-wise aggregation by imitating heterogeneous neighborhood structures, so as to recover informative and complete node attributes through capturing fine-grained graph contextual patterns. Based on the completed features, GFD-GC further introduces a confidence-aware supervised contrastive learning strategy, which augments scarce labeled fraud nodes with high-confidence pseudo-fraud nodes. 
In summary, the main contributions of this paper are as follows:
\begin{itemize}
    \item We propose a novel graph fraud detection framework, termed \textbf{GFD-GC}, to jointly address incomplete node attributes and extreme class imbalance.
    \item We design a grouped attribute completion module that captures heterogeneous neighborhood semantics for more effective feature recovery.
    \item We develop a confidence-aware contrastive learning strategy that leverages high-confidence pseudo-fraud nodes to improve fraud representation learning under scarce labels.
    \item Extensive experiments on real-world datasets show that \textbf{GFD-GC} consistently outperforms state-of-the-art baselines.
\end{itemize}

\section{RELATED WORK}
\label{sec:rw}

\subsection{Graph Fraud Detection}
Graph fraud detection aims to identify fraudulent nodes from relational data such as users, accounts, devices, and transactions. Since such dependencies can be naturally modeled as graphs, Graph Neural Networks (GNNs) have become a dominant solution due to their ability to capture structural and contextual information~\cite{gcn,graphsage,gat,gin}. Based on general GNN backbones, many fraud-oriented methods have been developed to address challenges such as camouflage, imbalance, heterophily, and anomalous graph patterns. For example, CARE-GNN~\cite{caregnn} improves robustness to camouflaged fraud via label-aware similarity and reinforcement-learning-based neighbor selection. PC-GNN~\cite{pcgnn} alleviates class imbalance through label-balanced sampling and selective aggregation. H2-FDetector~\cite{h2fdetector} models both homophilic and heterophilic relations, while BWGNN~\cite{bwgnn} studies fraud detection from a spectral perspective with Beta wavelets. Although these methods achieve promising results, they mainly focus on designing stronger fraud detectors under complete feature settings. Recent studies further enhance graph representations by modeling collaborative, modularized, high-order, and dynamic graph patterns \cite{gtcn2026,ncigcn2025,mgcn2025,highordergcn2026,sgddyg2025}. In real-world scenarios, however, node attributes are often incomplete, which can substantially degrade representation quality and detection performance. 

\subsection{Learning with Incomplete Node Attributes}
Learning on graphs with incomplete node attributes has attracted increasing attention in recent years. A straightforward solution is to fill missing entries with zeros, means, or other predefined values before training. However, such strategies ignore graph structure and cannot effectively recover informative features. To address this issue, recent studies explore graph-aware completion or joint representation learning. Graph-based incomplete-data models improve representation learning under incomplete high-dimensional features through graph pooling, robust autoencoding, and multi-metric latent feature analysis \cite{glcp2025,orae2025,mmlf2024}. Nevertheless, they are mainly designed for general representation learning and do not explicitly consider the heterogeneous neighborhood structures commonly observed in fraud graphs. As a result, their recovered features may still be insufficient for downstream fraud detection.

\subsection{Contrastive Learning for Imbalanced Graph Representation Learning}
Contrastive learning improves representation robustness and discriminability by maximizing consistency between related samples. Supervised contrastive learning~\cite{supcon} enhances intra-class compactness and inter-class separability, while contrastive distillation has shown effectiveness in fraud-related detection scenarios \cite{qinetheft2026}. This is desirable for graph fraud detection, where fraudulent nodes are scarce and difficult to distinguish. However, directly applying supervised contrastive learning is still limited by the shortage of labeled fraud nodes.

\begin{figure*}[t]
    \centering
    \includegraphics[width=0.95\textwidth]{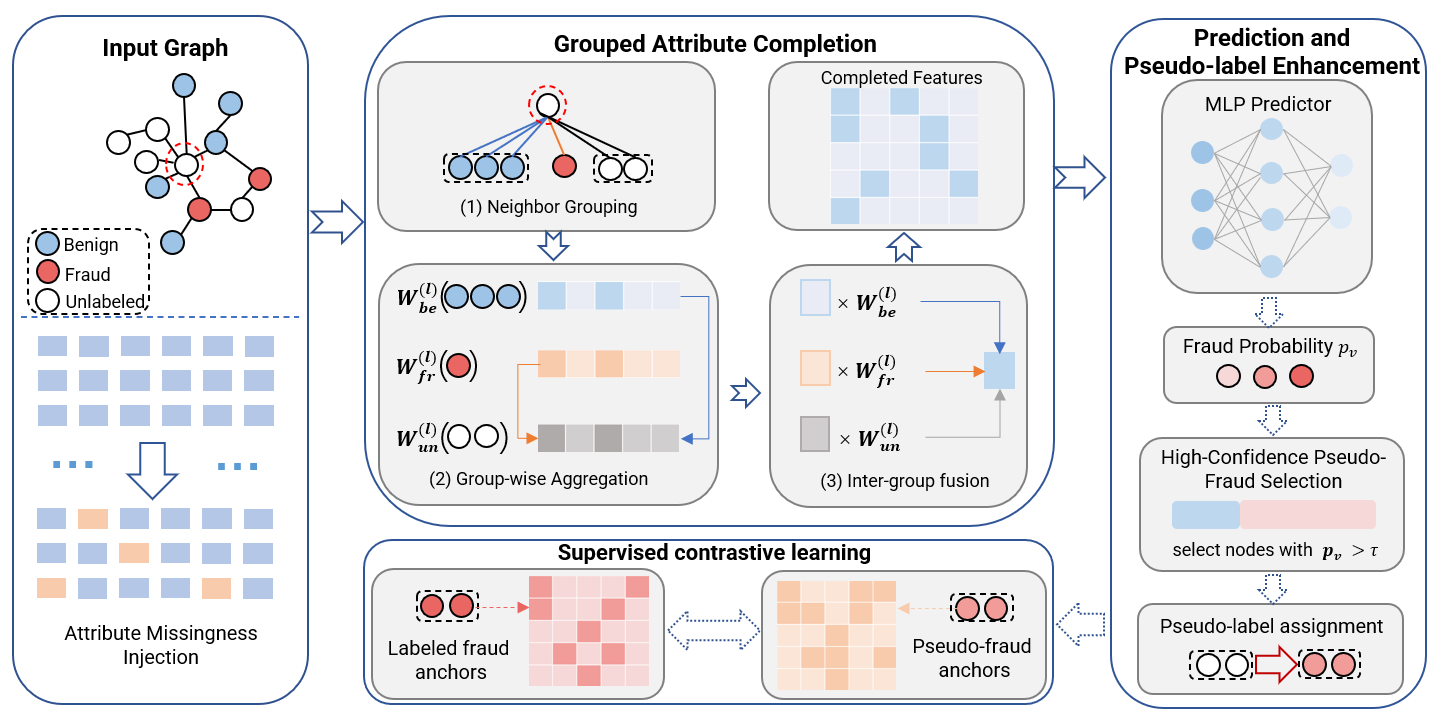}
    \caption{Overall architecture of the proposed GFD-GC framework, which consists of grouped attribute completion and confidence-aware supervised contrastive learning for graph fraud detection.}
    \label{fig:pipeline}
\end{figure*}

\section{METHOD}
\label{sec:method}

\subsection{Problem Definition}
Let $G=(V,E)$ be an undirected graph with $|V|=N$ nodes and feature matrix $\mathbf{X}\in\mathbb{R}^{N\times F}$, where each node $v\in V$ is associated with a feature vector $\mathbf{x}_v\in\mathbb{R}^{F}$. A subset of training nodes $V_{\mathrm{tr}}^L\subseteq V$ is labeled and visible during training, with binary labels $y_v\in\{0,1\}$ indicating fraud and non-fraud nodes. The goal is to learn a classifier $f_\theta$ that predicts the fraud probability of each node.

To model incomplete node attributes, we introduce an observation mask $\mathbf{M}\in\{0,1\}^{N\times F}$, where $M_{v,f}=1$ indicates that the $f$-th feature of node $v$ is observed, and $M_{v,f}=0$ otherwise. The corrupted feature matrix is defined as
\begin{equation}
\tilde{\mathbf{X}}=\mathbf{M}\odot \mathbf{X},
\end{equation}
where $\odot$ denotes element-wise multiplication. Given $G$, $\tilde{\mathbf{X}}$, and the labels on $V_{\mathrm{tr}}^L$, the task is to recover missing attributes and detect fraud jointly.

\subsection{Framework Overview}
Figure~\ref{fig:pipeline} shows \textbf{GFD-GC}, which contains grouped attribute completion and confidence-aware supervised contrastive learning. Given corrupted features and graph structure, GFD-GC partitions neighbors into benign, fraud, and unknown groups according to training-label visibility, performs group-specific aggregation, and selects high-confidence pseudo-fraud nodes to enhance supervised contrastive learning.

\subsection{Grouped Attribute Completion}
\label{subsec:completion}
The grouped attribute completion module recovers missing node attributes from neighborhood context by explicitly modeling heterogeneous neighborhood information.

\subsubsection{Mask-aware node encoding}
For each node $v$, we concatenate its corrupted feature vector $\tilde{\mathbf{x}}_v$ with its mask vector $\mathbf{m}_v$:
\begin{equation}
\mathbf{r}_v=[\tilde{\mathbf{x}}_v \Vert \mathbf{m}_v]\in\mathbb{R}^{2F},
\end{equation}
where $\mathbf{m}_v$ is the $v$-th row of $\mathbf{M}$. The concatenated input is projected into a hidden representation:
\begin{equation}
\mathbf{h}_v=\sigma(\mathbf{W}_e\mathbf{r}_v+\mathbf{b}_e)\in\mathbb{R}^{H},
\end{equation}
where $\mathbf{W}_e\in\mathbb{R}^{H\times 2F}$ and $\mathbf{b}_e\in\mathbb{R}^{H}$ are learnable parameters, and $\sigma(\cdot)$ is a nonlinear activation function.

\subsubsection{Group-wise aggregation over heterogeneous neighbors}
Let $\mathcal{N}(v)=\{u\mid (u,v)\in E\}$ denote the 1-hop neighbors of node $v$. Since only training labels are visible during model optimization, we divide the neighbors of each node into three groups according to their label visibility:
\begin{equation}
\begin{aligned}
\mathcal{N}_{be}(v)&=\{u\in \mathcal{N}(v)\cap V_{\mathrm{tr}}^L \mid y_u=0\},\\
\mathcal{N}_{fr}(v)&=\{u\in \mathcal{N}(v)\cap V_{\mathrm{tr}}^L \mid y_u=1\},\\
\mathcal{N}_{un}(v)&=\mathcal{N}(v)\setminus
\left(\mathcal{N}_{be}(v)\cup \mathcal{N}_{fr}(v)\right).
\end{aligned}
\end{equation}
Here, $\mathcal{N}_{be}(v)$, $\mathcal{N}_{fr}(v)$, and $\mathcal{N}_{un}(v)$ denote benign, fraud, and label-unknown neighbors, respectively. Validation and test nodes are always included in the unknown group during training to avoid label leakage.

For each group $g\in\{be,fr,un\}$, we apply a group-specific transformation matrix $\mathbf{W}_{g}^{(l)}$ at the $l$-th aggregation layer. The group-wise representation is computed as
\begin{equation}
\mathbf{s}_{v,g}^{(l)}
=
\sum_{u\in\mathcal{N}_{g}(v)}
\alpha_{v,u}^{g,l}
\mathbf{W}_{g}^{(l)}
\mathbf{h}_{u}^{(l)},
\end{equation}
where $\alpha_{v,u}^{g,l}$ is the attention weight of neighbor $u$ within group $g$. It is defined as
\begin{equation}
\alpha_{v,u}^{g,l}
=
\mathrm{softmax}_{u\in\mathcal{N}_{g}(v)}
\left(
\phi_g^{(l)}
\left(
[\mathbf{h}_{v}^{(l)}\Vert \mathbf{h}_{u}^{(l)}]
\right)
\right),
\end{equation}
where $\phi_g^{(l)}(\cdot)$ is a learnable group-specific scoring function.

\subsubsection{Inter-group fusion}
After obtaining group-wise neighborhood representations, we estimate the importance of each group via an inter-group attention mechanism:
\begin{equation}
\beta_{v,g}^{(l)}
=
\mathrm{softmax}_{g\in\{be,fr,un\}}
\big(
\eta^{(l)}([\mathbf{h}_{v}^{(l)}\Vert \mathbf{s}_{v,g}^{(l)}])
\big),
\end{equation}
where $\eta^{(l)}(\cdot)$ is a learnable function. The fused neighborhood representation is
\begin{equation}
\mathbf{s}_v^{(l)}
=
\sum_{g\in\{be,fr,un\}}
\beta_{v,g}^{(l)}
\mathbf{s}_{v,g}^{(l)}.
\end{equation}

\subsubsection{Feature completion}
Given the node encoding $\mathbf{h}_v$ and fused neighborhood representation $\mathbf{s}_v$, we generate a completed feature proposal:
\begin{equation}
\bar{\mathbf{x}}_v=\mathbf{W}_r[\mathbf{h}_v\Vert \mathbf{s}_v]+\mathbf{b}_r\in\mathbb{R}^{F},
\end{equation}
where $\mathbf{W}_r$ and $\mathbf{b}_r$ are learnable parameters. To preserve observed entries, only missing dimensions are replaced:
\begin{equation}
\hat{\mathbf{x}}_v=\mathbf{m}_v\odot \tilde{\mathbf{x}}_v+(\mathbf{1}-\mathbf{m}_v)\odot \bar{\mathbf{x}}_v.
\label{eq:completion}
\end{equation}
The completion module is optimized jointly with fraud detection without a separate reconstruction loss.

\subsection{MLP Fraud Predictor}
Based on the completed features $\hat{\mathbf{X}}$, we employ an MLP predictor to generate node embeddings and classification logits:
\begin{equation}
\mathbf{z}_v=f_\theta^{(\mathrm{pen})}(\hat{\mathbf{x}}_v)\in\mathbb{R}^{D},\qquad
\boldsymbol{\ell}_v=f_\theta^{(\mathrm{cls})}(\mathbf{z}_v)\in\mathbb{R}^{2},
\end{equation}
where $\mathbf{z}_v$ denotes the penultimate-layer embedding and $\boldsymbol{\ell}_v$ denotes the output logits. The predicted fraud probability of node $v$ is
\begin{equation}
p_v=\mathrm{softmax}(\boldsymbol{\ell}_v)[1].
\end{equation}

The classification branch is optimized on labeled training nodes using the cross-entropy loss:
\begin{equation}
\mathcal{L}_{cls}=\frac{1}{|V_{\mathrm{tr}}^L|}\sum_{v\in V_{\mathrm{tr}}^L}\mathrm{CE}(\boldsymbol{\ell}_v,y_v),
\end{equation}
where $\mathrm{CE}(\cdot)$ denotes the cross-entropy function.

\subsection{Confidence-Aware Supervised Contrastive Learning}
\label{subsec:contrast}
Cross-entropy supervision alone is often insufficient under extreme class imbalance, as the scarcity of labeled fraud nodes limits the discriminability of fraud representations. To address this issue, we introduce a confidence-aware supervised contrastive learning strategy.

\subsubsection{Fraud-oriented contrastive space}
We first normalize node embeddings:
\begin{equation}
\tilde{\mathbf{z}}_v=\frac{\mathbf{z}_v}{\|\mathbf{z}_v\|_2}.
\end{equation}
Let
\begin{equation}
P=\{v\in V_{\mathrm{tr}}^L \mid y_v=1\}
\end{equation}
denote the set of labeled fraud nodes. Since $|P|$ is typically very small, directly applying supervised contrastive learning on $P$ alone provides limited positive supervision.

\subsubsection{High-confidence pseudo-fraud augmentation}
To enrich fraud-oriented supervision, we augment labeled fraud nodes with pseudo-fraud nodes selected from label-unknown data:
\begin{equation}
\hat{P}=\{v\in V\setminus V_{\mathrm{tr}}^L \mid p_v>\tau\},
\end{equation}
where $\tau\in(0,1)$ is a confidence threshold. Nodes in $\hat{P}$ are treated as high-confidence pseudo-fraud nodes. These pseudo labels are used only in the contrastive branch and do not affect the supervised classification loss.

\subsubsection{Contrastive objective}
The anchor set is defined as
\begin{equation}
A=P\cup\hat{P}.
\end{equation}
For each anchor $i\in A$, its positive set is
\begin{equation}
Pos(i)=A\setminus\{i\}.
\end{equation}
We adopt an InfoNCE-style supervised contrastive objective:
\begin{equation}
\mathcal{L}_{con}=
-\frac{1}{|A|}\sum_{i\in A}\frac{1}{|Pos(i)|}\sum_{j\in Pos(i)}
\log
\frac{\exp(\tilde{\mathbf{z}}_i^\top \tilde{\mathbf{z}}_j/T)}
{\sum_{k\neq i}\exp(\tilde{\mathbf{z}}_i^\top \tilde{\mathbf{z}}_k/T)},
\end{equation}
where $T$ is the temperature parameter. This objective pulls fraud-related nodes closer while pushing them away from non-fraud nodes, thereby improving the compactness and separability of fraud representations.

\subsection{Joint Training Objective}
The final objective combines fraud classification and confidence-aware contrastive learning:
\begin{equation}
\mathcal{L}=\mathcal{L}_{cls}+\lambda_{con}\mathcal{L}_{con},
\end{equation}
where $\lambda_{con}$ is a trade-off coefficient. The two terms are jointly optimized, allowing the grouped completion module to recover features that are directly beneficial to downstream fraud detection.

\section{EXPERIMENTS}
\label{sec:exp}

\subsection{Experimental Setup}

\paragraph{Datasets and splits}
We evaluate the proposed method on two widely used graph fraud detection benchmarks, \textbf{Amazon} and \textbf{Yelp} \cite{reviewnet,yelpfraud,caregnn,pcgnn}. Each node is associated with an attribute vector and a binary fraud label. Following prior work \cite{caregnn,pcgnn,h2fdetector,bwgnn}, we adopt a \emph{stratified} split to preserve the class ratio. Specifically, $40\%$ of labeled nodes are used for training, while the remaining labeled nodes are divided into validation and test sets at a ratio of $1{:}2$, resulting in $20\%$ validation and $40\%$ test nodes. For Amazon, unlabeled nodes are removed from both training and evaluation.

\paragraph{Missingness injection}
To evaluate robustness under incomplete node attributes, we inject synthetic feature missingness with rate
$\rho \in \{20\%,30\%,40\%,50\%,60\%,70\%\}$.
For each setting, a binary observation mask $\mathbf{M}$ is sampled over the whole graph and the same missingness pattern is applied to the training, validation, and test sets. Unless otherwise stated, we use \textbf{40\% missingness} for the main comparison and report all missing rates for robustness and ablation analysis.

\paragraph{Baselines}
We compare GFD-GC with representative graph fraud detection and graph anomaly detection baselines, including
GCN~\cite{gcn}, GAT~\cite{gat}, GIN~\cite{gin}, GraphSAGE~\cite{graphsage}, CARE-GNN~\cite{caregnn}, BWGNN~\cite{bwgnn}, SparseGAD~\cite{sparsegad}, UniGAD~\cite{unigad}, and U-A2GAD~\cite{ua2gad}.
All methods are evaluated under the same split and missingness setting.

\paragraph{Evaluation metrics}
We report \textbf{Macro-F1}, \textbf{Macro-Recall}, and \textbf{AUC-ROC}. Macro-F1 and Macro-Recall emphasize performance on the minority fraud class under severe imbalance, while AUC-ROC evaluates the overall ranking quality of fraud predictions.

\paragraph{Implementation details}
The grouped attribute completion module and the fraud predictor are trained jointly using Adam. To alleviate class imbalance, benign nodes are under-sampled when constructing training batches. Model selection is based on validation AUC with early stopping, and the results are reported from the best checkpoint. Key hyperparameters, including the missing rate $\rho$, the confidence threshold $\tau$ for pseudo-fraud selection, and the contrastive loss weight $\lambda_{con}$, are tuned on the validation set.

\begin{table*}[t]
\centering
\caption{Performance comparison of \textbf{GFD-GC} and baseline methods on graph fraud detection under \textbf{40\%} missing node attributes. Best results are in \textbf{bold} and second-best results are \underline{underlined}. The last column reports the relative change of \textbf{GFD-GC} compared with the strongest baseline for each metric.}
\label{tab:main40_grouped}
\setlength{\tabcolsep}{3.2pt}
\renewcommand{\arraystretch}{1.15}
\resizebox{\textwidth}{!}{
\begin{tabular}{cc|cccc|cc|ccc|c|c}
\toprule
{Dataset} & {Metric}
& \multicolumn{4}{c|}{GCN-based Models}
& \multicolumn{2}{c|}{Fraud GNNs}
& \multicolumn{3}{c|}{Anomaly/SSL Models}
& \multicolumn{1}{c|}{Ours}
& \multirow{2}{*}{Gain} \\
\cmidrule(lr){3-6}\cmidrule(lr){7-8}\cmidrule(lr){9-11}\cmidrule(lr){12-12}
& & GCN & GAT & GIN & GraphSAGE
& CARE-GNN & BWGNN
& SparseGAD & UniGAD & U-A2GAD
& \textbf{GFD-GC}
&  \\
\midrule
\multirow{3}{*}{Amazon}
& F1     & 0.5542 & 0.6422 & 0.5645 & 0.5450 & 0.7439 & 0.7548 & 0.7496 & 0.7058 & \underline{0.7624} & \textbf{0.7840} & +2.83\% \\
& Recall & 0.7467 & 0.7364 & 0.7083 & 0.6853 & 0.8235 & 0.7053 & 0.7328 & \textbf{0.8485} & 0.7333 & \underline{0.8390} & -1.12\% \\
& AUC    & 0.8421 & 0.8055 & 0.7785 & 0.7442 & 0.8770 & 0.8556 & 0.8523 & 0.9050 & \underline{0.9073} & \textbf{0.9093} & +0.22\% \\
\midrule
\multirow{3}{*}{Yelp}
& F1     & 0.5297 & 0.4810 & 0.5419 & 0.4701 & 0.4963 & 0.5348 & \underline{0.5439} & 0.5047 & 0.5323 & \textbf{0.5633} & +3.57\% \\
& Recall & 0.5302 & 0.5071 & 0.5401 & 0.5021 & 0.6019 & 0.5476 & 0.5465 & \underline{0.6034} & 0.5455 & \textbf{0.6327} & +4.86\% \\
& AUC    & 0.5520 & 0.5262 & 0.5618 & 0.5194 & \underline{0.6452} & 0.5942 & 0.5909 & 0.6445 & 0.6447 & \textbf{0.6689} & +3.67\% \\
\bottomrule
\end{tabular}
}
\end{table*}

\begin{figure*}[htbp]
    \centering
    \includegraphics[width=\textwidth]{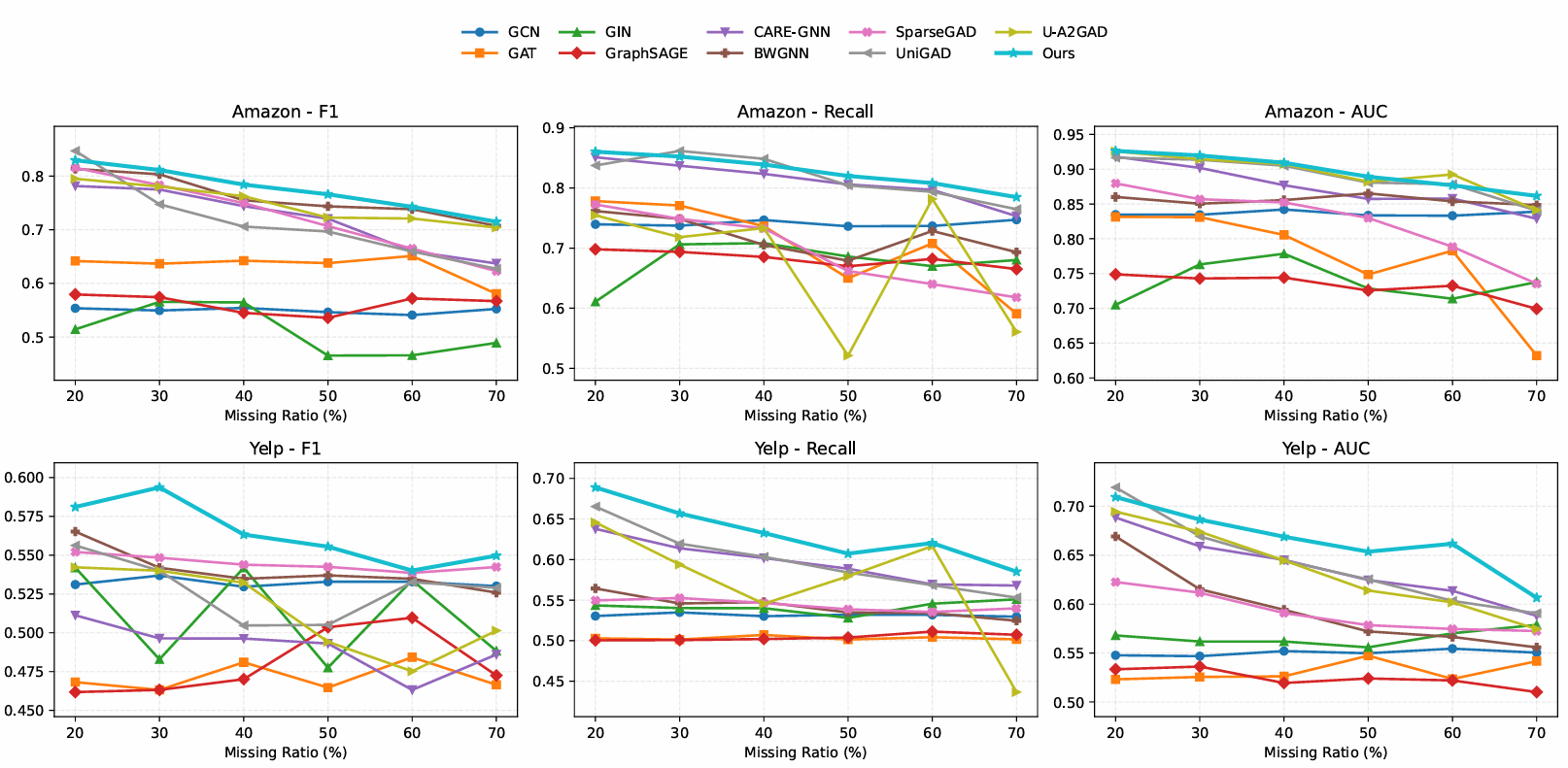}
    \caption{Performance comparison of different methods under varying missing ratios on Amazon and Yelp datasets in terms of F1, Recall, and AUC.}
    \label{fig:missing-ratio-results}
\end{figure*}

\begin{figure*}[htbp]
    \centering
    \includegraphics[width=\textwidth]{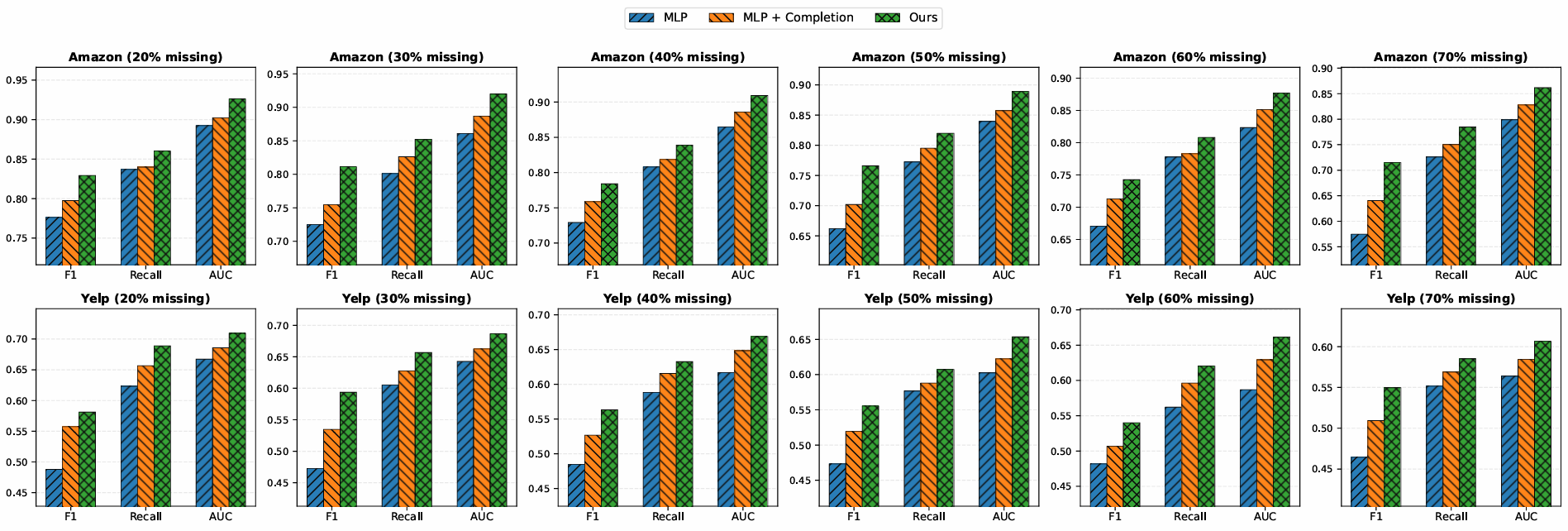}
    \caption{Ablation study under different missing ratios on Amazon and Yelp datasets.}
    \label{fig:ablation-missing-ratio}
\end{figure*}

\subsection{Main Results under 40\% Missingness}
\label{subsec:main40}

Table~\ref{tab:main40_grouped} reports the results under \textbf{40\%} missing features. Overall, GFD-GC achieves the best \textbf{Macro-F1} and \textbf{AUC} on both datasets, demonstrating superior fraud detection performance under incomplete node attributes.

On \textbf{Amazon}, GFD-GC improves Macro-F1 from $0.7624$ with U-A2GAD to $0.7840$, and slightly improves AUC from $0.9073$ to $0.9093$. Although UniGAD achieves the highest recall, our method remains competitive ($0.8390$ vs.\ $0.8485$). These results indicate that the proposed framework improves the overall discrimination of fraud nodes while maintaining strong sensitivity.

On \textbf{Yelp}, GFD-GC achieves the best performance on all three metrics. Compared with the strongest baseline, it improves Macro-F1 by $3.57\%$, Macro-Recall by $4.86\%$, and AUC by $3.67\%$. This verifies that the proposed grouped attribute completion and confidence-aware contrastive learning are particularly effective in challenging fraud graphs with corrupted attributes and scarce fraud labels.

\begin{table}[t]
\centering
\caption{Ablation results under \textbf{40\%} missing features. Best in \textbf{bold}, second best underlined.}
\label{tab:ablation40}
\setlength{\tabcolsep}{4pt}
\renewcommand{\arraystretch}{1.15}
\begin{tabular}{cc|cc|c}
\toprule
{Dataset} & {Metric}
& \multicolumn{2}{c|}{Baselines}
& \multicolumn{1}{c}{Ours} \\
\cmidrule(lr){3-4}\cmidrule(lr){5-5}
& & MLP & MLP+Comp & Ours \\
\midrule
{Amazon}
& F1     & 0.7292 & \underline{0.7587} & \textbf{0.7840} \\
& Recall & 0.8080 & \underline{0.8185} & \textbf{0.8390} \\
& AUC    & 0.8646 & \underline{0.8856} & \textbf{0.9093} \\
\midrule
{Yelp}
& F1     & 0.4846 & \underline{0.5268} & \textbf{0.5633} \\
& Recall & 0.5885 & \underline{0.6157} & \textbf{0.6327} \\
& AUC    & 0.6171 & \underline{0.6485} & \textbf{0.6689} \\
\bottomrule
\end{tabular}
\end{table}

\subsection{Ablation Study}
\label{subsec:ablation}

We study three variants: \textbf{MLP}, \textbf{MLP+Completion}, and \textbf{Ours}. 
\textbf{MLP} directly uses corrupted features with missing entries filled by zeros.
\textbf{MLP+Completion} adds the proposed grouped attribute completion module but removes the contrastive loss by setting $\lambda_{con}=0$.
\textbf{Ours} is the full model with both grouped attribute completion and confidence-aware supervised contrastive learning.

Table~\ref{tab:ablation40} reports the ablation results under \textbf{40\%} missingness, and Fig.~\ref{fig:ablation-missing-ratio} presents the trends across all missing rates from $20\%$ to $70\%$. First, adding the completion module consistently outperforms plain MLP on both datasets, showing that group-wise aggregation over heterogeneous neighborhoods can effectively recover informative node features under incomplete attributes. Second, incorporating confidence-aware supervised contrastive learning brings further gains on all metrics, indicating that high-confidence pseudo-fraud augmentation provides additional fraud-oriented supervision and improves the compactness and separability of fraud representations.

More importantly, the full model remains consistently superior across different missing rates. As shown in Fig.~\ref{fig:ablation-missing-ratio}, \textbf{Ours} outperforms both ablated variants on Amazon and Yelp in almost all settings, and the performance gap becomes more evident under heavier missingness.

\subsection{Robustness to Different Missing Rates}
\label{subsec:robust}

To further evaluate robustness, we vary the missing rate from $20\%$ to $70\%$. Figure~\ref{fig:missing-ratio-results} shows the performance of all compared methods on Amazon and Yelp.

On \textbf{Amazon}, GFD-GC achieves the best Macro-F1 across all missing rates and remains among the strongest methods in Recall and AUC, demonstrating stable robustness under feature corruption.

On \textbf{Yelp}, GFD-GC consistently achieves the best Macro-F1 and Macro-Recall and remains competitive in AUC across all missing rates, confirming the effectiveness of grouped completion and confidence-aware contrastive supervision.

\section{CONCLUSIONS}
This paper studies graph fraud detection under two practical challenges, namely incomplete node attributes and extreme class imbalance. To address these issues, we propose \textbf{GFD-GC}, a novel framework that integrates grouped attribute completion with confidence-aware supervised contrastive learning. Specifically, GFD-GC imitates heterogeneous neighborhood structures to perform group-wise aggregation, thereby recovering informative node features by capturing fine-grained graph contextual patterns. It further augments scarce labeled fraud nodes with high-confidence pseudo-fraud nodes, which improves the compactness of fraud representations and their separability from non-fraud nodes. Extensive experiments on two real-world datasets demonstrate that GFD-GC consistently outperforms state-of-the-art baselines and remains robust under different levels of attribute missingness. These results verify the effectiveness of the proposed framework and highlight its potential for real-world fraud detection scenarios.


\begin{thebibliography}{95}

\bibitem{qinetheft2026}
W.~Qin, Y.~Ding, and X.~Luo, ``A Robust Approach to Electricity Theft Detection via Tensor Representation-Driven Contrastive Distillation,'' \emph{IEEE Transactions on Industrial Informatics}, doi: 10.1109/TII.2026.3659333, 2026.

\bibitem{crypto2025}
X.~Liao, H.~Wu, T.~He, and X.~Luo, ``A Proximal-ADMM-incorporated Nonnegative Latent-Factorization-of-Tensors Model for Representing Dynamic Cryptocurrency Transaction Network,'' \emph{IEEE Transactions on Systems, Man, and Cybernetics: Systems}, vol.~55, no.~11, pp.~8387--8401, 2025.

\bibitem{gcn}
T.~N. Kipf and M.~Welling, ``Semi-Supervised Classification with Graph Convolutional Networks,'' in \emph{Proc. ICLR}, 2017.

\bibitem{graphsage}
W.~Hamilton, Z.~Ying, and J.~Leskovec, ``Inductive Representation Learning on Large Graphs,'' in \emph{Proc. NeurIPS}, 2017.

\bibitem{gat}
P.~Veli\v{c}kovi\'{c}, G.~Cucurull, A.~Casanova, A.~Romero, P.~Li\`{o}, and Y.~Bengio, ``Graph Attention Networks,'' in \emph{Proc. ICLR}, 2018.

\bibitem{gin}
K.~Xu, W.~Hu, J.~Leskovec, and S.~Jegelka, ``How Powerful are Graph Neural Networks?,'' in \emph{Proc. ICLR}, 2019.

\bibitem{ncigcn2025}
Y.~Yuan, Y.~Wang, and X.~Luo, ``A Node-Collaboration-Informed Graph Convolutional Network for Highly Accurate Representation to Undirected Weighted Graph,'' \emph{IEEE Transactions on Neural Networks and Learning Systems}, vol.~36, no.~6, pp.~11507--11519, 2025.

\bibitem{mgcn2025}
T.~He, Z.~Duan, and X.~Luo, ``Modularized Graph Convolutional Network,'' \emph{IEEE/CAA Journal of Automatica Sinica}, doi: 10.1109/JAS.2025.125336, 2025.

\bibitem{gtcn2026}
L.~Wang, Y.~Yuan, and X.~Luo, ``Graph Tensor Convolutional Network,'' \emph{IEEE Transactions on Systems, Man, and Cybernetics: Systems}, doi: 10.1109/TSMC.2026.3655418, 2026.

\bibitem{reviewnet}
S.~Rayana and L.~Akoglu, ``Collective Opinion Spam Detection: Bridging Review Networks and Metadata,'' in \emph{Proc. KDD}, 2015.

\bibitem{yelpfraud}
J.~J. McAuley and J.~Leskovec, ``From Amateurs to Connoisseurs: Modeling the Evolution of User Expertise through Online Reviews,'' in \emph{Proc. WWW}, 2013.

\bibitem{caregnn}
Y.~Dou, Z.~Liu, L.~Sun, Y.~Deng, H.~Peng, and P.~S. Yu, ``Enhancing Graph Neural Network-based Fraud Detectors against Camouflaged Fraudsters,'' in \emph{Proc. CIKM}, 2020.

\bibitem{pcgnn}
Y.~Liu, X.~Ao, F.~Feng, Q.~Yang, X.~He, K.~Li, and Q.~He, ``Pick and Choose: A GNN-based Imbalanced Learning Approach for Fraud Detection,'' in \emph{Proc. WWW}, 2021.

\bibitem{sgddyg2025}
M.~Han, L.~Wang, Y.~Yuan, and X.~Luo, ``SGD-DyG: Self-Reliant Global Dependency Apprehending on Dynamic Graphs,'' in \emph{Proc. ACM SIGKDD Conference on Knowledge Discovery and Data Mining}, 2025, pp.~802--813.

\bibitem{glcp2025}
F.~Bi, T.~He, Y.-S.~Ong, and X.~Luo, ``Graph Linear Convolution Pooling for Learning in Incomplete High-Dimensional Data,'' \emph{IEEE Transactions on Knowledge and Data Engineering}, vol.~37, no.~4, pp.~1838--1852, 2025.

\bibitem{orae2025}
D.~Wu, Y.~Hu, K.~Liu, J.~Li, X.~Wang, S.~Deng, N.~Zheng, and X.~Luo, ``An Outlier-Resilient Autoencoder for Representing High-Dimensional and Incomplete Data,'' \emph{IEEE Transactions on Emerging Topics in Computational Intelligence}, vol.~9, no.~2, pp.~1379--1391, 2025.

\bibitem{mmlf2024}
D.~Wu, P.~Zhang, Y.~He, and X.~Luo, ``MMLF: Multi-Metric Latent Feature Analysis for High-Dimensional and Incomplete Data,'' \emph{IEEE Transactions on Services Computing}, vol.~17, no.~2, pp.~575--588, 2024.

\bibitem{supcon}
P.~Khosla, P.~Teterwak, C.~Wang, A.~Sarna, Y.~Tian, P.~Isola, A.~Maschinot, C.~Liu, and D.~Krishnan, ``Supervised Contrastive Learning,'' in \emph{Proc. NeurIPS}, 2020.

\bibitem{h2fdetector}
F.~Shi, Y.~Cao, Y.~Shang, Y.~Zhou, C.~Zhou, and J.~Wu, ``H2-FDetector: A GNN-based Fraud Detector with Homophilic and Heterophilic Connections,'' in \emph{Proc. WWW}, 2022.

\bibitem{bwgnn}
J.~Tang, J.~Li, Z.~Gao, and J.~Li, ``Rethinking Graph Neural Networks for Anomaly Detection,'' in \emph{Proc. ICML}, 2022.

\bibitem{highordergcn2026}
L.~Wang, Y.~Yuan, and X.~Luo, ``Advanced High-Order Graph Convolutional Networks with Assorted Time-Frequency Transforms,'' \emph{IEEE/CAA Journal of Automatica Sinica}, vol.~13, no.~2, pp.~394--408, 2026.

\bibitem{sparsegad}
Z.~Gong, G.~Wang, Y.~Sun, Q.~Liu, Y.~Ning, H.~Xiong, and J.~Peng, ``Beyond Homophily: Robust Graph Anomaly Detection via Neural Sparsification,'' in \emph{Proc. IJCAI}, 2023.

\bibitem{unigad}
Y.~Lin, J.~Tang, C.~Zi, H.~V. Zhao, Y.~Yao, and J.~Li, ``UniGAD: Unifying Multi-level Graph Anomaly Detection,'' in \emph{Proc. NeurIPS}, 2024.

\bibitem{ua2gad}
Y.~Li, G.~Zang, C.~Song, and X.~Yuan, ``A Universal Adaptive Algorithm for Graph Anomaly Detection,'' \emph{Information Processing \& Management}, vol.~62, no.~1, Art.~103905, 2025.




\bibitem{ekflfa2026}
Y.~Yuan, S.~Wang, H.~Zhou, L.~Wang, and X.~Luo, ``A Novel Approach to Temporal QoS Estimation via Extended Kalman Filter-Incorporated Latent Feature Analysis,'' \emph{IEEE Transactions on Services Computing}, doi: 10.1109/TSC.2026.3697552, 2026.

\bibitem{apidnlf2026}
J.~Li, Y.~Yuan, T.~He, and X.~Luo, ``Adaptive PID-Incorporated Nonnegative Latent Factor Analysis,'' \emph{IEEE Transactions on Systems, Man, and Cybernetics: Systems}, doi: 10.1109/TSMC.2026.3678292, 2026.

\bibitem{lerpid2025}
J.~Li, Y.~Yuan, and X.~Luo, ``Learning Error Refinement in Stochastic Gradient Descent-based Latent Factor Analysis via Diversified PID Controllers,'' \emph{IEEE Transactions on Emerging Topics in Computational Intelligence}, vol.~9, no.~5, pp.~3582--3597, 2025.

\bibitem{pinlf2025}
Y.~Yuan, S.~Lu, and X.~Luo, ``A Proportional Integral Controller-Enhanced Non-negative Latent Factor Analysis Model,'' \emph{IEEE/CAA Journal of Automatica Sinica}, vol.~12, no.~6, pp.~1246--1259, 2025.

\bibitem{gta2t2025}
L.~Wang, K.~Liu, and Y.~Yuan, ``GT-A2T: Graph Tensor Alliance Attention Network,'' \emph{IEEE/CAA Journal of Automatica Sinica}, vol.~12, no.~10, pp.~2165--2167, 2025.

\bibitem{fuzzypid2024}
Y.~Yuan, J.~Li, and X.~Luo, ``A Fuzzy PID-Incorporated Stochastic Gradient Descent Algorithm for Fast and Accurate Latent Factor Analysis,'' \emph{IEEE Transactions on Fuzzy Systems}, vol.~32, no.~7, pp.~4049--4061, 2024.

\bibitem{adnlfa2024}
Y.~Yuan, X.~Luo, and M.~Zhou, ``Adaptive Divergence-based Non-negative Latent Factor Analysis of High-Dimensional and Incomplete Matrices from Industrial Applications,'' \emph{IEEE Transactions on Emerging Topics in Computational Intelligence}, vol.~8, no.~2, pp.~1209--1222, 2024.

\bibitem{kflfa2023}
Y.~Yuan, X.~Luo, M.~Shang, and Z.~Wang, ``A Kalman-Filter-Incorporated Latent Factor Analysis Model for Temporally Dynamic Sparse Data,'' \emph{IEEE Transactions on Cybernetics}, vol.~53, no.~9, pp.~5788--5801, 2023.

\bibitem{adnlfm2023}
Y.~Yuan, R.~Wang, G.~Yuan, and X.~Luo, ``An Adaptive Divergence-based Non-negative Latent Factor Model,'' \emph{IEEE Transactions on Systems, Man, and Cybernetics: Systems}, vol.~53, no.~10, pp.~6475--6487, 2023.

\bibitem{mrlfm2022}
Y.~Yuan, Q.~He, X.~Luo, and M.~Shang, ``A Multilayered-and-Randomized Latent Factor Model for High-Dimensional and Sparse Matrices,'' \emph{IEEE Transactions on Big Data}, vol.~8, no.~3, pp.~784--794, 2022.

\bibitem{gfnlf2020}
Y.~Yuan, X.~Luo, M.~Shang, and D.~Wu, ``A Generalized and Fast-converging Non-negative Latent Factor Model for Predicting User Preferences in Recommender Systems,'' in \emph{Proc. WWW}, 2020, pp.~498--507.

\bibitem{kfwebqos2020}
Y.~Yuan, M.~Shang, and X.~Luo, ``Temporal Web Service QoS Prediction via Kalman Filter-Incorporated Dynamic Latent Factor Analysis,'' in \emph{Proc. ECAI}, 2020, pp.~561--568.

\bibitem{nghfl2026}
D.~Wu, S.~Li, Y.~He, X.~Luo, and X.~Gao, ``Non-Gradient Hash Factor Learning for High-Dimensional and Incomplete Data Representation Learning,'' \emph{IEEE Transactions on Pattern Analysis and Machine Intelligence}, doi: 10.1109/TPAMI.2026.3653780, 2026.

\bibitem{modetucker2025}
H.~Wu, Q.~Wang, X.~Luo, and Z.~Wang, ``Learning Accurate Representation to Nonstandard Tensors via a Mode-Aware Tucker Network,'' \emph{IEEE Transactions on Knowledge and Data Engineering}, vol.~37, no.~12, pp.~7272--7285, 2025.

\bibitem{dsrpso2026}
C.~Lyu, Z.~Ma, X.~Luo, and Y.~Shi, ``Dynamic Stochastic Reorientation Particle Swarm Optimization for Adaptive Latent Factor Analysis in High-Dimensional Sparse Matrices,'' \emph{IEEE Transactions on Knowledge and Data Engineering}, vol.~38, no.~1, pp.~222--234, 2026.

\bibitem{tlroc2026}
Y.~He and X.~Luo, ``Tensor Low-Rank Orthogonal Compression for Convolutional Neural Networks,'' \emph{IEEE/CAA Journal of Automatica Sinica}, vol.~13, no.~1, pp.~227--229, 2026.

\bibitem{dgmixer2025}
F.~Bi, T.~He, Y.-S.~Ong, and X.~Luo, ``Discovering Spatio-Temporal-Individual Coupled Features from Nonstandard Tensors-A Novel Dynamic Graph Mixer Approach,'' \emph{IEEE Transactions on Neural Networks and Learning Systems}, vol.~36, no.~11, pp.~19834--19848, 2025.

\bibitem{stgnntensor2025}
F.~Bi, T.~He, and X.~Luo, ``Spatiotemporal Graph Neural Network-Incorporated Latent Factorization of Tensors for Dynamic QoS Estimation,'' \emph{IEEE/CAA Journal of Automatica Sinica}, doi: 10.1109/JAS.2025.125750, 2025.

\bibitem{antf2025}
P.~Tang, X.~Luo, and J.~Woodcock, ``Auto-Encoding Neural Tucker Factorization,'' \emph{IEEE Transactions on Knowledge and Data Engineering}, vol.~37, no.~10, pp.~5795--5807, 2025.

\bibitem{gatwostep2025}
C.~Lyu, J.~Cheng, X.~Luo, and Y.~Shi, ``Genetic Algorithm-based Two-Step Optimization for Precise Latent Factor Analysis,'' \emph{IEEE Transactions on Neural Networks and Learning Systems}, doi: 10.1109/TNNLS.2025.3631465, 2025.

\bibitem{ncsac2026}
L.~Lin, Q.~Li, M.~Qiao, Z.~Wang, J.~Zhao, R.-H.~Li, X.~Luo, and T.~Jia, ``NCSAC: Effective Neural Community Search via Attribute-augmented Conductance,'' \emph{IEEE Transactions on Knowledge and Data Engineering}, vol.~38, no.~2, pp.~1221--1235, 2026.

\bibitem{mmae2025}
D.~Wu, C.~Liang, Y.~He, Y.~Qiao, and X.~Luo, ``Multi Metric Autoencoder for Representing High-Dimensional and Incomplete Data,'' \emph{IEEE Transactions on Systems, Man, and Cybernetics: Systems}, doi: 10.1109/TSMC.2025.3646863, 2025.

\bibitem{hifreview2025}
Q.~Hu, H.~Wu, and X.~Luo, ``A Comprehensive Review of Parallel Optimization Algorithms for High-Dimensional and Incomplete Matrix Factorization,'' \emph{IEEE/CAA Journal of Automatica Sinica}, vol.~12, no.~12, pp.~2399--2426, 2025.

\bibitem{snrtlft2026}
X.~Xu, M.~Lin, Z.~Xu, and X.~Luo, ``A Sampling-Neighborhood-Regularized Latent Factorization of Tensor for Dynamic QoS Estimation,'' \emph{IEEE Transactions on Network and Service Management}, vol.~23, pp.~1707--1722, 2026.

\bibitem{tccn2025}
X.~Liao, H.~Wu, and X.~Luo, ``A Novel Tensor Causal Convolution Network Model for Highly-Accurate Representation to Spatio-Temporal Data,'' \emph{IEEE Transactions on Automation Science and Engineering}, vol.~22, pp.~19525--19537, 2025.

\bibitem{cbnlf2025}
Q.~Wang, H.~Wu, and X.~Luo, ``A Convolution Bias-Incorporated Nonnegative Latent Factorization of Tensors Model for Accurate Representation Learning to Dynamic Directed Graphs,'' \emph{IEEE Transactions on Systems, Man, and Cybernetics: Systems}, vol.~55, no.~12, pp.~8902--8914, 2025.

\bibitem{amnlft2025}
X.~Xu, M.~Lin, Z.~Xu, and X.~Luo, ``Attention-Mechanism-Based Neural Latent-Factorization-of-Tensors Mode,'' \emph{ACM Transactions on Knowledge Discovery from Data}, vol.~19, no.~4, pp.~1--27, 2025.

\bibitem{kdmil2025}
C.~Li, P.~Huang, J.~Qin, and X.~Luo, ``Knowledge-driven Multiple Instance Learning with Hierarchical Cluster-incorporated Aware Filtering for Larynx Pathological Grading,'' \emph{IEEE Journal of Biomedical and Health Informatics}, doi: 10.1109/JBHI.2025.3609838, 2025.

\bibitem{labc2025}
Y.~Yang, L.~Hu, G.~Li, D.~Li, P.~Hu, and X.~Luo, ``Link-based Attributed Graph Clustering via Approximate Generative Bayesian Learning,'' \emph{IEEE Transactions on Systems, Man, and Cybernetics: Systems}, vol.~55, no.~8, pp.~5730--5743, 2025.

\bibitem{fmvpci2025}
Y.~Yang, L.~Hu, G.~Li, D.~Li, P.~Hu, and X.~Luo, ``Fmvpci: A Multi-View Fusion Neural Network for Identifying Protein Complex via Fuzzy Clustering,'' \emph{IEEE Transactions on Systems, Man, and Cybernetics: Systems}, vol.~55, no.~9, pp.~6189--6202, 2025.

\bibitem{biasnlt2025}
X.~Xu, M.~Lin, X.~Luo, and Z.~Xu, ``An Adaptively Bias-Extended Non-negative Latent Factorization of Tensors Model for Accurately Representing the Dynamic QoS Data,'' \emph{IEEE Transactions on Services Computing}, vol.~18, no.~2, pp.~603--617, 2025.

\bibitem{mirna2025}
M.-Y.~Wu, P.~Hu, Z.-H.~You, J.~Zhang, L.~Hu, and X.~Luo, ``Graph-Based Prediction of miRNA-Drug Associations with Multisource Information and Metapath Enhancement Matrices,'' \emph{IEEE Journal of Biomedical and Health Informatics}, doi: 10.1109/JBHI.2025.3558303, 2025.

\bibitem{masntf2025}
Y.~Hou, P.~Tang, and X.~Luo, ``Multi-Aspect Self-Attending Neural Tucker Factorization for Spatiotemporal Representation Learning,'' \emph{IEEE/CAA Journal of Automatica Sinica}, doi: 10.1109/JAS.2025.125723, 2025.

\bibitem{nnlft2026}
W.~Li, M.~Lin, X.~Xu, L.~Lin, Z.~Xu, and X.~Luo, ``Neural Non-Negative Latent Factorization of Tensors Model with Acceleration and Unconstraint,'' \emph{IEEE Transactions on Systems, Man, and Cybernetics: Systems}, vol.~56, no.~1, pp.~164--178, 2026.

\bibitem{fmoe2026}
X.~Deng, P.~Hu, T.~Herget, F.~Tan, X.~Zhu, J.~Zhang, Y.-a.~Huang, L.~Hu, Z.~You, and X.~Luo, ``Fuzzy Mixture-of-Experts Aggregation for Organoid Identification with Multi-Scale State Space Features,'' \emph{IEEE Transactions on Fuzzy Systems}, vol.~34, no.~1, pp.~324--335, 2026.

\bibitem{mcdgnn2026}
J.~Gou, Y.~Cheng, B.~Ma, L.~Du, X.~Luo, and Z.~Yi, ``Multi-Scale Collaborative Distillation Graph Neural Networks for Session-Based Recommendation,'' \emph{IEEE Transactions on Services Computing}, vol.~19, no.~1, pp.~504--517, 2026.

\bibitem{sentiment2025}
J.~Liu, X.~Li, M.~Lin, and X.~Luo, ``A Scalable Multi-Channel Sentiment Analysis Model with Enhanced Semantic Understanding and Redundancy Reduction,'' \emph{IEEE Transactions on Computational Social Systems}, doi: 10.1109/TCSS.2025.3619188, 2025.

\bibitem{emotion2025}
Z.~Luo, X.~Jin, Y.~Luo, Q.~Zhou, and X.~Luo, ``Analysis of Students' Positive Emotion and Smile Intensity Using Sequence-Relative Key-Frame Labeling and Deep-Asymmetric Convolutional Neural Network,'' \emph{IEEE/CAA Journal of Automatica Sinica}, vol.~12, no.~4, pp.~806--820, 2025.

\bibitem{spsa2025}
Z.~He, M.~Lin, X.~Luo, and Z.~Xu, ``Structure-Preserved Self-Attention for Fusion Image Information in Multiple Color Spaces,'' \emph{IEEE Transactions on Neural Networks and Learning Systems}, vol.~36, no.~7, pp.~13021--13035, 2025.

\bibitem{batterylife2025}
M.~Chen, L.~Tao, J.~Lou, and X.~Luo, ``Latent Factorization of Tensors Incorporated Battery Cycle Life Prediction,'' \emph{IEEE/CAA Journal of Automatica Sinica}, vol.~12, no.~3, pp.~633--635, 2025.

\bibitem{rllfa2025}
D.~Wu, Z.~Li, Z.~Yu, Y.~He, and X.~Luo, ``Robust Low-rank Latent Feature Analysis for Spatio-Temporal Signal Recovery,'' \emph{IEEE Transactions on Neural Networks and Learning Systems}, vol.~36, no.~2, pp.~2829--2842, 2025.

\bibitem{ntf2025}
P.~Tang and X.~Luo, ``Neural Tucker Factorization,'' \emph{IEEE/CAA Journal of Automatica Sinica}, vol.~12, no.~2, pp.~475--477, 2025.

\bibitem{trafficimputation2025}
H.~Yang, M.~Lin, H.~Chen, X.~Luo, and Z.~Xu, ``Latent Factor Analysis Model with Temporal Regularized Constraint for Road Traffic Data Imputation,'' \emph{IEEE Transactions on Intelligent Transportation Systems}, vol.~26, no.~1, pp.~724--741, 2025.

\bibitem{anytime2024}
X.~Liao, K.~Hoang, and X.~Luo, ``Local Search-based Anytime Algorithms for Continuous Distributed Constraint Optimization Problems,'' \emph{IEEE/CAA Journal of Automatica Sinica}, vol.~12, no.~1, pp.~1--3, 2024.

\bibitem{fgreg2024}
H.~Wu, Y.~Qiao, and X.~Luo, ``A Fine-Grained Regularization Scheme for Nonnegative Latent Factorization of High-Dimensional and Incomplete Tensors,'' \emph{IEEE Transactions on Services Computing}, vol.~17, no.~6, pp.~3006--3021, 2024.

\bibitem{admmnlf2024}
Y.~Zhong, K.~Liu, S.~Gao, and X.~Luo, ``Alternating-Direction-Method of Multipliers-based Adaptive Nonnegative Latent Factor Analysis,'' \emph{IEEE Transactions on Emerging Topics in Computing}, vol.~8, no.~5, pp.~3544--3558, 2024.

\bibitem{kdaerial2024}
N.~Zeng, X.~Li, P.~Wu, H.~Li, and X.~Luo, ``A Novel Tensor Decomposition-based Efficient Detector for Low-altitude Aerial Objects with Knowledge Distillation Scheme,'' \emph{IEEE/CAA Journal of Automatica Sinica}, vol.~11, no.~2, pp.~487--501, 2024.

\bibitem{robotarmsystem2024}
Z.~Li, S.~Li, and X.~Luo, ``A Novel Machine Learning System for Industrial Robot Arm Calibration,'' \emph{IEEE Transactions on Circuits and Systems II: Express Briefs}, vol.~71, no.~4, pp.~2364--2368, 2024.

\bibitem{apsgd2024}
W.~Qin, X.~Luo, and M.~Zhou, ``Adaptively-accelerated Parallel Stochastic Gradient Descent for High-Dimensional and Incomplete Data Representation Learning,'' \emph{IEEE Transactions on Big Data}, vol.~10, no.~1, pp.~92--107, 2024.

\bibitem{deepwalk2014}
B.~Perozzi, R.~Al-Rfou, and S.~Skiena, ``DeepWalk: Online Learning of Social Representations,'' in \emph{Proc. KDD}, 2014, pp.~701--710.

\bibitem{node2vec2016}
A.~Grover and J.~Leskovec, ``node2vec: Scalable Feature Learning for Networks,'' in \emph{Proc. KDD}, 2016, pp.~855--864.

\bibitem{line2015}
J.~Tang, M.~Qu, M.~Wang, M.~Zhang, J.~Yan, and Q.~Mei, ``LINE: Large-scale Information Network Embedding,'' in \emph{Proc. WWW}, 2015, pp.~1067--1077.

\bibitem{chebnet2016}
M.~Defferrard, X.~Bresson, and P.~Vandergheynst, ``Convolutional Neural Networks on Graphs with Fast Localized Spectral Filtering,'' in \emph{Proc. NeurIPS}, 2016, pp.~3844--3852.

\bibitem{sgc2019}
F.~Wu, A.~Souza, T.~Zhang, C.~Fifty, T.~Yu, and K.~Q. Weinberger, ``Simplifying Graph Convolutional Networks,'' in \emph{Proc. ICML}, 2019, pp.~6861--6871.

\bibitem{appnp2019}
J.~Klicpera, A.~Bojchevski, and S.~G{\"u}nnemann, ``Predict then Propagate: Graph Neural Networks meet Personalized PageRank,'' in \emph{Proc. ICLR}, 2019.

\bibitem{jknet2018}
K.~Xu, C.~Li, Y.~Tian, T.~Sonobe, K.-i.~Kawarabayashi, and S.~Jegelka, ``Representation Learning on Graphs with Jumping Knowledge Networks,'' in \emph{Proc. ICML}, 2018, pp.~5453--5462.

\bibitem{vgae2016}
T.~N. Kipf and M.~Welling, ``Variational Graph Auto-Encoders,'' in \emph{Proc. NeurIPS Workshop on Bayesian Deep Learning}, 2016.

\bibitem{dgi2019}
P.~Veli{\v{c}}kovi{\'c}, W.~Fedus, W.~L. Hamilton, P.~Li{\`o}, Y.~Bengio, and R.~D. Hjelm, ``Deep Graph Infomax,'' in \emph{Proc. ICLR}, 2019.

\bibitem{graphcl2020}
Y.~You, T.~Chen, Y.~Sui, T.~Chen, Z.~Wang, and Y.~Shen, ``Graph Contrastive Learning with Augmentations,'' in \emph{Proc. NeurIPS}, 2020.

\bibitem{mvgrl2020}
K.~Hassani and A.~H. Khasahmadi, ``Contrastive Multi-View Representation Learning on Graphs,'' in \emph{Proc. ICML}, 2020, pp.~4116--4126.

\bibitem{grace2020}
Y.~Zhu, Y.~Xu, F.~Yu, Q.~Liu, S.~Wu, and L.~Wang, ``Deep Graph Contrastive Representation Learning,'' in \emph{Proc. ICML Workshop on Graph Representation Learning and Beyond}, 2020.

\bibitem{gca2021}
Y.~Zhu, Y.~Xu, F.~Yu, Q.~Liu, S.~Wu, and L.~Wang, ``Graph Contrastive Learning with Adaptive Augmentation,'' in \emph{Proc. WWW}, 2021, pp.~2069--2080.

\bibitem{simclr2020}
T.~Chen, S.~Kornblith, M.~Norouzi, and G.~Hinton, ``A Simple Framework for Contrastive Learning of Visual Representations,'' in \emph{Proc. ICML}, 2020, pp.~1597--1607.

\bibitem{moco2020}
K.~He, H.~Fan, Y.~Wu, S.~Xie, and R.~Girshick, ``Momentum Contrast for Unsupervised Visual Representation Learning,'' in \emph{Proc. CVPR}, 2020, pp.~9729--9738.

\bibitem{graphsmote2021}
T.~Zhao, X.~Zhang, and S.~Wang, ``GraphSMOTE: Imbalanced Node Classification on Graphs with Graph Neural Networks,'' in \emph{Proc. WSDM}, 2021, pp.~833--841.

\bibitem{gcnmf2021}
H.~Taguchi, X.~Liu, and T.~Murata, ``Graph Convolutional Networks for Graphs Containing Missing Features,'' \emph{Future Generation Computer Systems}, vol.~117, pp.~155--168, 2021.

\bibitem{sat2022}
X.~Chen, S.~Chen, J.~Yao, H.~Zheng, Y.~Zhang, and I.~W. Tsang, ``Learning on Attribute-Missing Graphs,'' \emph{IEEE Transactions on Pattern Analysis and Machine Intelligence}, vol.~44, no.~2, pp.~740--757, 2022.

\bibitem{featureprop2022}
E.~Rossi, H.~Kenlay, M.~I. Gorinova, B.~P. Chamberlain, X.~Dong, and M.~Bronstein, ``On the Unreasonable Effectiveness of Feature Propagation in Learning on Graphs with Missing Node Features,'' in \emph{Proc. LoG}, 2022.

\bibitem{dominant2019}
K.~Ding, J.~Li, R.~Bhanushali, and H.~Liu, ``Deep Anomaly Detection on Attributed Networks,'' in \emph{Proc. SDM}, 2019, pp.~594--602.

\bibitem{anomalydae2020}
H.~Fan, F.~Zhang, and Z.~Li, ``AnomalyDAE: Dual Autoencoder for Anomaly Detection on Attributed Networks,'' in \emph{Proc. ICASSP}, 2020, pp.~5685--5689.

\bibitem{cola2022}
Y.~Liu, Z.~Li, S.~Pan, C.~Gong, C.~Zhou, and G.~Karypis, ``Anomaly Detection on Attributed Networks via Contrastive Self-Supervised Learning,'' \emph{IEEE Transactions on Neural Networks and Learning Systems}, vol.~33, no.~6, pp.~2378--2392, 2022.

\end{thebibliography}
\end{document}